\documentclass[10pt,twocolumn,letterpaper]{article}

\usepackage{cvpr}
\usepackage{times}
\usepackage{epsfig}
\usepackage{graphicx}
\usepackage{amsmath}
\usepackage{amssymb}
\usepackage{multicol}
\usepackage[dvipsnames]{xcolor}

\begin{document}

\title{Radial Distortion in Face Images: Detection and Impact}

\author{Wassim Kabbani \quad Tristan Le Pessot* \quad Kiran Raja \quad Raghavendra Ramachandra, Christoph Busch \\
NTNU, Gjøvik, Norway \quad ENSICAEN, Caen, France* \\
{\tt\small \{wassim.h.kabbani; kiran.raja; raghavendra.ramachandra; christoph.busch \} @ntnu.no} \\
{\tt\small tristan.le-pessot@orange.fr}
}

\maketitle
\thispagestyle{empty}




\begin{abstract}
    Acquiring face images of sufficiently high quality is important for online ID and travel document issuance applications using face recognition systems (FRS). Low-quality, manipulated (intentionally or unintentionally), or distorted images degrade the FRS performance and facilitate documents' misuse. Securing quality for enrolment images, especially in the unsupervised self-enrolment scenario via a smartphone, becomes important to assure FRS performance. In this work, we focus on the less studied area of radial distortion (a.k.a., the fish-eye effect) in face images and its impact on FRS performance. We introduce an effective radial distortion detection model that can detect and flag radial distortion in the enrolment scenario. We formalize the detection model as a face image quality assessment (FIQA) algorithm and provide a careful inspection of the effect of radial distortion on FRS performance. Evaluation results show excellent detection results for the proposed models, and the study on the impact on FRS uncovers valuable insights into how to best use these models in operational systems.
\end{abstract}


\section{Introduction}
\label{sec:intro}

\begin{figure}[t]
     \begin{center}
     \begin{subfigure}[b]{0.30\linewidth}
         \centering
         \includegraphics[width=\linewidth]{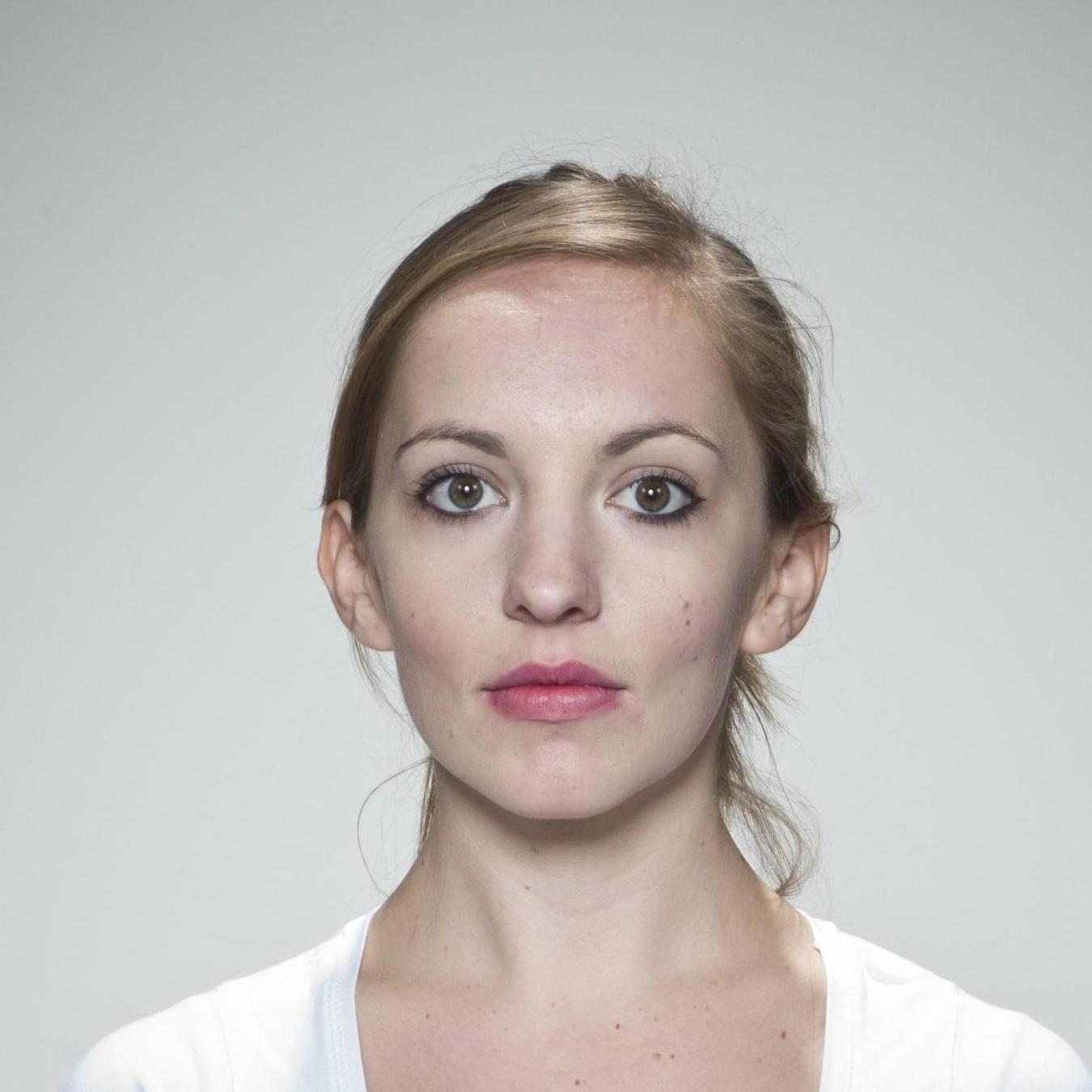}
         \caption{Original}
         \label{fig:frll-1-original}
     \end{subfigure}
     \hfill
     \begin{subfigure}[b]{0.30\linewidth}
         \centering
         \includegraphics[width=\linewidth]{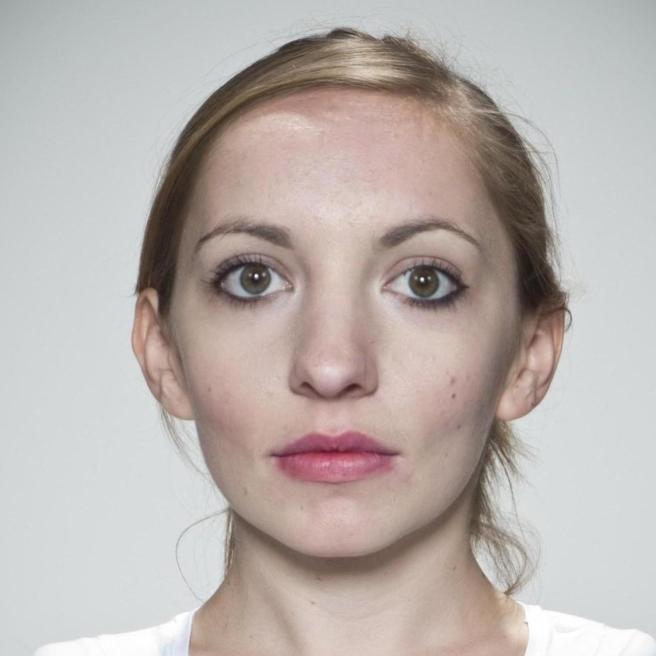}
         \caption{DM\textsubscript{$\lambda = 0.4$}}
         \label{fig:frll-1-dm4}
     \end{subfigure}
     \hfill
     \begin{subfigure}[b]{0.30\linewidth}
         \centering
         \includegraphics[width=\linewidth]{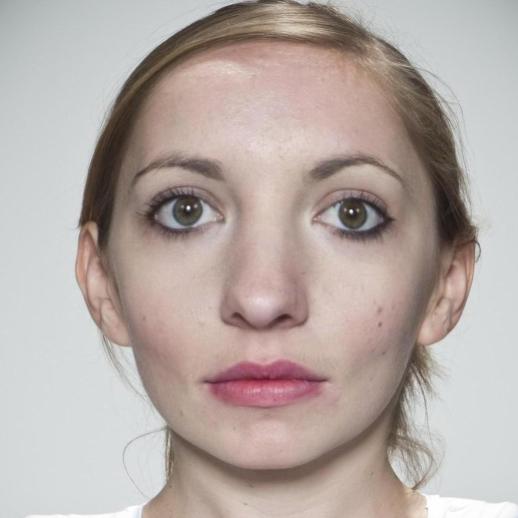}
         \caption{DM\textsubscript{$\lambda = 0.9$}}
         \label{fig:frll-1-dm9}
     \end{subfigure}
     \begin{subfigure}[b]{0.30\linewidth}
         \centering
         \includegraphics[width=\linewidth]{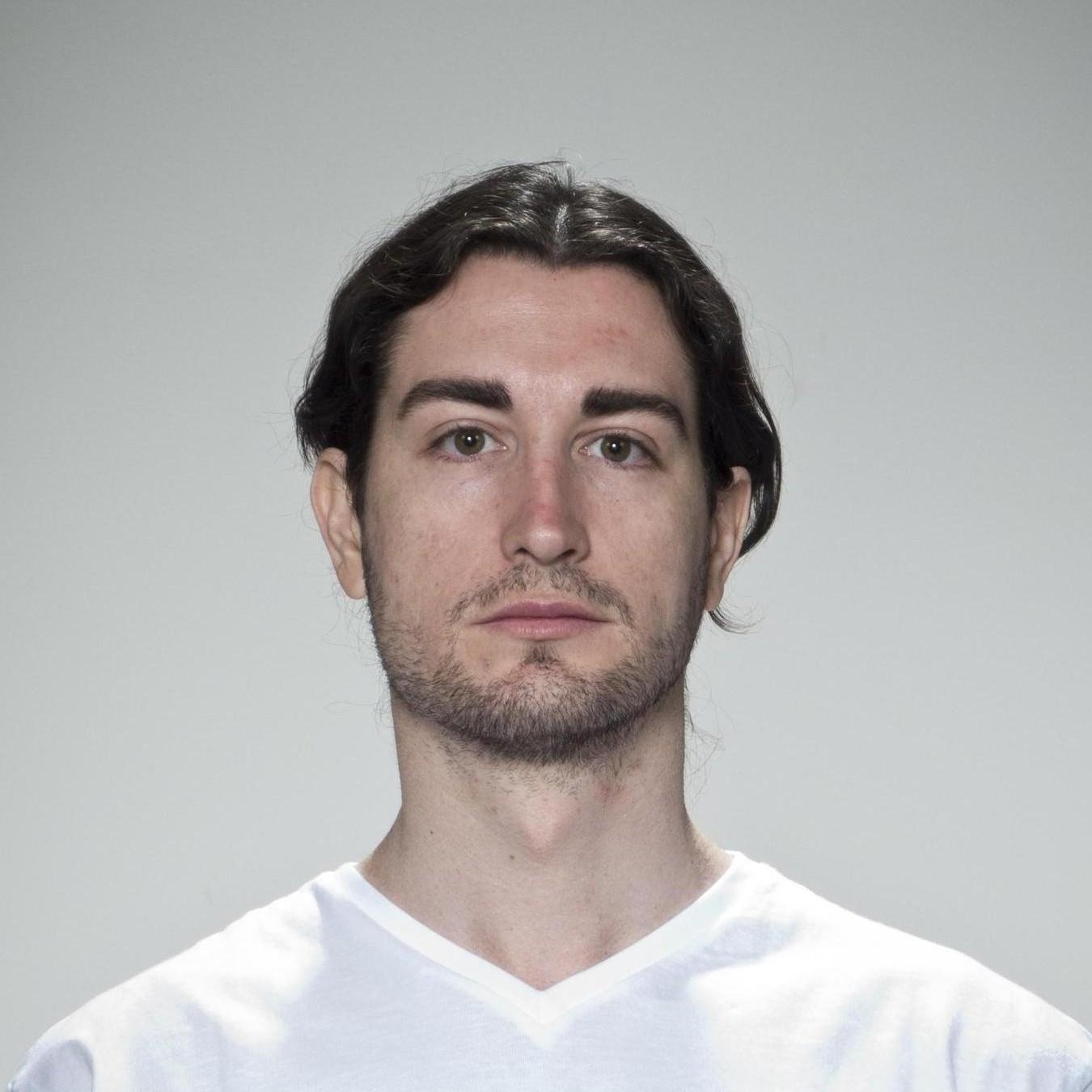}
         \caption{Original}
         \label{fig:frll-2-original}
     \end{subfigure}
     \hfill
     \begin{subfigure}[b]{0.30\linewidth}
         \centering
         \includegraphics[width=\linewidth]{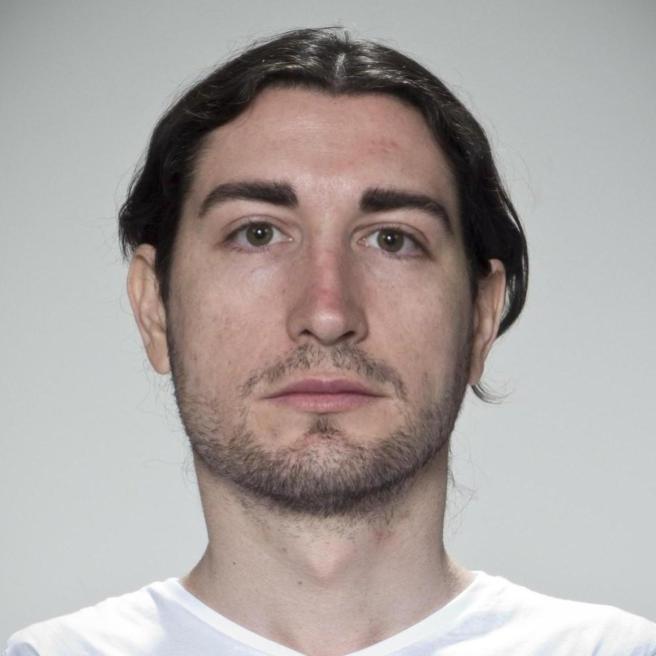}
         \caption{DM\textsubscript{$\lambda = 0.4$}}
         \label{fig:frll-2-dm4}
     \end{subfigure}
     \hfill
     \begin{subfigure}[b]{0.30\linewidth}
         \centering
         \includegraphics[width=\linewidth]{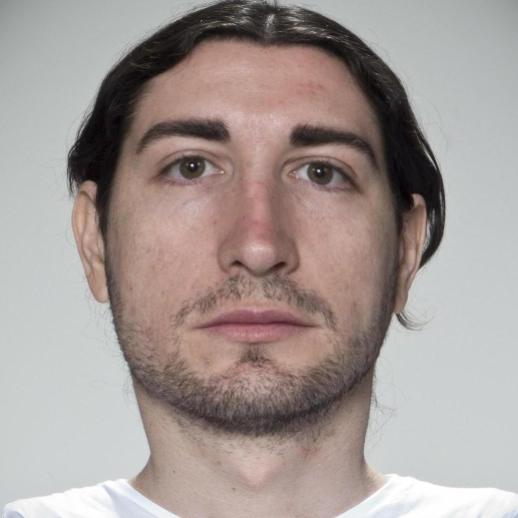}
         \caption{DM\textsubscript{$\lambda = 0.9$}}
         \label{fig:frll-2-dm9}
     \end{subfigure}
     \end{center}
    \caption{Radial distortion effect in face images. The left column shows the original, undistorted images. The middle column shows images with a mild level of distortion. The right column shows the images with a more pronounced level of distortion. Distortion is introduced synthetically using the Division Model (DM) but with different values for the distortion coefficient $\lambda$. Images from the FRLL dataset \cite{DeBruine2021}.}
    \label{fig:frll-samples}
\end{figure}

Face recognition systems (FRS) are being deployed at a number of airports, border crossings, and security gates. They have become a cornerstone of our security architecture to verify identity. Online ID and travel document issuance applications that are mobile/smartphone-compatible are being rolled out in many countries \footnote{https://www.ireland.ie/en/dfa/passports/passport-online/} \footnote{https://www.cbp.gov/travel/us-citizens/mobile-passport-control}. These applications allow individuals to self-enroll by uploading an existing face photo or taking a selfie, which will then be used on the issued document as a reference image to verify the identity of the individual in a later encounter. The reliability of these systems and the reference data stored therein is therefore of great importance. FRS needs to not only perform well but also be quality-aware and immune to various kinds of attacks. Many studies have shown that low-quality images have a negative impact on FRS performance \cite{ISO-IEC-29794-5-DIS-FaceQuality-240129, Schlett-FIQA-LiteratureSurvey-CSUR-2021, Boutros-CR-FIQA-CVPR-2023}, and further, they have a negative impact on detecting presentation attacks (also morphing) \cite{Aravena-2022, Biying-2022}. Robust methods that can examine the quality of the incoming images are therefore needed to flag any non-complaint or suspected image, making sure the obtained images are as representative as possible of the individual to whom they belong. According to the ISO/IEC 29794-5 standard \cite{ISO-IEC-29794-5-DIS-FaceQuality-240129}, the quality assessment of a face image belongs to capture-related and subject-related quality components. Capture-related quality components such as background uniformity, illumination uniformity, under/over exposure, and natural color, and subject-related quality components such as eyes open, mouth closed, head size, and pose have dedicated metrics to measure them. Radial distortion is yet another capture-related quality component emerging due to the optical properties of the camera. Commonly known as the fish-eye effect, it distorts the captured image, creating a hemispherical (panoramic) effect. The disturbance effect on the image and the perception of the represented face are shown in the example images in Figure \ref{fig:frll-samples}. Such an image distortion is an effect caused by user interaction that is not compliant with the capture process policy (for example, the smartphone is too close to the subject) and is not necessarily an attack. The standard ISO/IEC 19794-5 \cite{ISO-IEC-19794-5-G2-FaceImage-110304} mandates the "fish eye" effect associated with wide-angle lenses to be absent. Further, the standard ISO/IEC 39794-5 \cite{ISO-IEC-39794-5-G3-FaceImage-191015} requires the maximum magnification distortion rate to be equal to or below 7\%. However, to the best of our knowledge, an algorithm to detect or estimate radial distortion in a face image is not yet contained in the draft quality testing version of the implementation-focused standard ISO/IEC 29794-5 \cite{ISO-IEC-29794-5-DIS-FaceQuality-240129}.

In the course of addressing the above-mentioned limitation, this paper primarily contributes with: (1) An effective radial distortion detection model that fits the unsupervised self-enrolment scenario and can flag radially distorted face images with high confidence. (2) A face image quality assessment (FIQA) algorithm based on the detection model that can produce a quality measure for the radial distortion quality component. (3) Insights into when to best run checks, such as radial distortion in an operational scenario. The secondary contributions are: (1) Formalizing the distortion transformations in a unified way, making them easier to use in future works. (2) Synthetic datasets for training and evaluating radial distortion models \footnote{The code and datasets will be made public along with the paper.}. The paper is organized as follows: Section \ref{sec:related} presents related work and preliminaries. Section \ref{sec:approach} introduces the proposed approach. Section \ref{sec:experiments} presents experiments and results. Sections \ref{sec:discussion} and \ref{sec:conclusion} present discussions and conclusions.

\section{Related Works}
\label{sec:related}

\subsection{Face Image Quality Assessment (FIQA)}

Face Image Quality Assessment (FIQA) refers to the process of assessing the quality of a face image and its utility for a face recognition system \cite{Schlett-FIQA-LiteratureSurvey-CSUR-2021}. As there are many aspects of the face image that could contribute to its quality, these aspects are being standardized in the ISO/IEC 29794-5 standard \cite{ISO-IEC-29794-5-DIS-FaceQuality-240129}, which uses the standardized term \textit{measure} to refer to all quality aspects of a face image. A FIQ measure can be either an end-to-end unified quality score that gives an overall assessment of the image or a quality component that assesses the quality of a specific aspect (or potentially a defect) of the face image, such as background uniformity, illumination uniformity, and natural color.

\subsubsection{Unified Quality Score} 

Many end-to-end methods that produce a unified quality score have been introduced. Hernandez-Ortega \etal introduced FaceQNet V0 \cite{hernandezortega2019faceqnet} and an optimized version later on in 2021 with FaceQNet V1 \cite{hernandezortega2021biometric}. Boutros \etal \cite{Boutros-CR-FIQA-CVPR-2023} presented a method for unified FIQA that depends on measuring the relative sample classifiability, which is measured based on the allocation of the training sample feature representation in angular space with respect to its class center and the nearest negative class center. They illustrate that there is a correlation between the face image quality and the sample's relative classifiability. 

\subsubsection{Quality Components} 

Several works addressed individual quality components. Hernandez-Ortega \cite{hernandez-ortega2021FaceQvec} proposed a framework to assess several quality components such as exposure, unnatural color, expression, pose, and background uniformity. The framework assesses image exposure by detecting if the image is too dark (or too bright) based on the mean pixel value. It uses a color-detector model to detect pixels with unnatural color and a pretrained CNN to detect non-neutral facial expression. Grimmer \etal \cite{NeutrEx-IJCB-2023} proposed an expression neutrality quality assessment method, \textit{NeutrEx}, that is based on the accumulated distances of a 3D face reconstruction to a neutral expression anchor \cite{EMOCA:CVPR:2021} \cite{DECA:Siggraph2021}. Yet, no work has addressed radial distortion in facial images and its utility as a face image quality component. The focus is mostly on distortion modeling for the purpose of rectification and not for measuring the quality/utility of the face image in FRS. In the mobile-based self-enrolment scenario, the aim is to obtain high-quality, non-distorted images that do not need rectification.


\subsection{Distortion Modeling and Rectification}

To detect or rectify radial distortion, it must first be modeled to estimate the distortion parameters and to be able to rectify the image. For modeling the distortion, a camera model that can describe lens distortion needs to be chosen, and then a method to estimate the distortion parameters can be applied \cite{Wu-CorrectionOfRD-SPIE-2017}. In this section, we highlight the most commonly used camera models for modeling radial distortion as well as relevant radial distortion rectification methods.

\subsubsection{Distortion Modeling}

A camera consists of an image plane and a lens, which produce a transformation between the 3D object space and the 2D image space. A perspective transformation cannot perfectly describe this transformation because of the distortion that occurs between points on the object and the location of their corresponding pixels on the image. This distortion can be modeled using a camera model. A camera model will provide an approximation of the transformation performed by the lens and the distortions that accompany it. Many camera models have been introduced, each with their own strengths and weaknesses; some are aimed at a specific type of camera lens, while others are more generic \cite{clarke1998development}. We provide a brief description of the two models used in this work.


\textbf{Division Model}: The Division Model (DM) \cite{fitzgibbon2001simultaneous} is a straightforward radial distortion model with advantageous algebraic qualities, the one-parameter model enables analytical reasoning for some issues that other distortion models can only tackle numerically. This distortion model has the characteristic that circles in the distorted image correspond to straight lines in the undistorted image. Since the radius of these circles can only be determined by using the distortion parameter and the center's location, the circles' centres serve as their complete descriptions.

\textbf{Kannala-Brandt Model}: The Kannala-Brandt model (KB) \cite{kannala2006generic} is a generic camera model suitable for fish-eye lens cameras as well as conventional and wide-angle cameras. The primary property of the KB model is that rather than characterizing radial distortion in terms of how far a point is from the image center (the radius), it characterizes distortion as a function of the incidence angle of the light passing through the lens. Following this, it suggests four variants of the model that better suit fish-eye lenses, namely, stereographic, equidistance, equisolid, and orthogonal projections.

\subsubsection{Distortion Rectification}

The majority of the works in the literature are focused on images of man-made environments, such as buildings and furnished rooms, and, to a lesser extent, relate to natural scenes or the faces of individuals. This is understandable given that most methods rely on visible straight lines in the image to perform the rectification. The rectification methods can be categorized into traditional handcrafted and end-to-end deep learning methods. The hand-crafted methods typically involve choosing a proper camera model to characterize the geometric distortions in the image, then estimating the distortion parameters of the camera by which the image was taken, and finally rectifying the image using the camera model and the estimated parameters \cite{bukhari2013automatic, fitzgibbon2001simultaneous, aleman2014automatic}. The deep learning methods, on the other hand, utilize the strong feature learning capacity of deep learning and do not explicitly detect geometric features in the fish-eye images but rather exploit the learned semantic representations to do the rectification. Most works create and utilize synthetic datasets for training and evaluating the models \cite{rong2017radial, yin2018fisheyerecnet, xue2020fisheye}.

\subsubsection{Distortion in Face Portraits} Few works present methods related to radial distortion in facial images. Fried \etal \cite{fried2016perspective} presented a method to modify the apparent relative pose and distance between the camera and the subject in a given single portrait photo. The method is not specifically aimed at correcting radial distortion but can help correct it as well. It can bring a distant camera closer to the subject, change the pose of the subject, create stereo pairs from an input portrait, and also correct artifacts related to radial distortion, such as large noses in selfies. Shih \etal \cite{shih2019distortion} introduced an algorithm to rectify faces without affecting other parts of the photo. Given a portrait as an input, the problem is formulated as an optimization problem to create a content-aware warping mesh that locally adapts to the stereographic projection on facial regions and evolves to the perspective projection over the background to efficiently correct distortion on facial regions for wide-angle portraits. Good rectification results on images taken by cameras ranging from 70° to 120° Field of View (FOV) angles, which include normal to ultra-wide-angle lenses found on most modern mobile phones is reported correspondingly \cite{shih2019distortion}.

\section{Proposed Approach}
\label{sec:approach}


Our proposed approach for radial distortion detection is based on deriving formal radial distortion transformations from two selected camera models, namely the Division Model (DM) and the Kannala-Brandt (KB) model. These two models offer the suitability to better model the radial distortion approximation and are able to create larger levels of distortion \cite{Wu-CorrectionOfRD-SPIE-2017}. We use the transformation models to create synthetic datasets suitable for training and evaluation. Then, we train and evaluate radial distortion detectors on both the synthetically created datasets and a newly collected real dataset. Finally, we formalize the detection model as a FIQA algorithm and use it to study the impact of radial distortion on FRS performance.

\subsection{Radial Distortion Transformations}
\label{sec:trans}

This section formalizes the radial distortion problem and transformations for two chosen camera models, addressing differences in descriptions, notation, and problem formulation across different works. Camera models primarily offer projections from object 3D point coordinates to image 2D pixel coordinates, but they can also be used to introduce or remove distortion in 2D images. Radial distortion models are described by a center of distortion $(x_c,y_c)$ and a function $\delta(r_d)$ that converts between distorted and undistorted radii, where $r_d$ is the Euclidean distance between $(x_c,y_c)$ and a point $(x_d,y_d)$ in the distorted image. $r_d$ is illustrated in Equation \ref{eq:euclidean}.

\begin{equation}
r_d = \sqrt{(x_d - x_c)^2 + (y_d - y_c)^2} 
\label{eq:euclidean}
\end{equation}

\subsubsection{Division Model}

According to the Division Model (DM) \cite{fitzgibbon2001simultaneous}, the $\delta(r_d)$ function is defined as in Equation \ref{eq:dm-theta} \cite{Wu-CorrectionOfRD-SPIE-2017}, where $\lambda_n$ are the distortion coefficients, and $n$ is the chosen number of coefficients to be used:

\begin{equation}
\delta(r_d) = 1 + \lambda_1r_d^2 + \lambda_2r_d^4 + ...
\label{eq:dm-theta}
\end{equation}

The relationship between the coordinates $(x_d, y_d)$ of a point in the distorted image and the coordinates $(x_u, y_u)$ of its corresponding location in the undistorted image is designated as follows \cite{Wu-CorrectionOfRD-SPIE-2017}:

\begin{equation}
x_u = x_c + \frac{x_d - x_c}{\delta(r_d)}
\label{eq:dm-x}
\end{equation}

\begin{equation}
y_u = y_c + \frac{y_d - y_c}{\delta(r_d)}
\label{eq:dm-y}
\end{equation}

In this work, we use only the first coefficient $\lambda_1$ because it induces the most distortion in the image compared to the following coefficients, and for most cameras, a single term is reported sufficient \cite{devernay2001straight, bukhari2013automatic}. We also consider the center of distortion to be the principal point of the image, i.e., $(x_c, y_c) = (0, 0)$ which is also a convention adopted in other works \cite{bukhari2013automatic}. This allows us to control the distortion level using a single distortion coefficient, $\lambda$ and simplifies equations \ref{eq:dm-x} and \ref{eq:dm-y} to the following:

\begin{equation}
x_u = \frac{x_d}{1 + \lambda r^2}
\label{eq:dm-x-sim}
\end{equation}

\begin{equation}
y_u = \frac{y_d}{1 + \lambda r^2}
\label{eq:dm-y-sim}
\end{equation}

\begin{figure*}[htp]
     \begin{center}
     \begin{subfigure}[b]{0.12\textwidth}
         \centering
         \includegraphics[width=\textwidth]{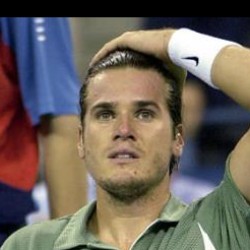}
         \caption{Original}
         \label{fig:lfw-1-original}
     \end{subfigure}
     \hfill
     \begin{subfigure}[b]{0.12\textwidth}
         \centering
         \includegraphics[width=\textwidth]{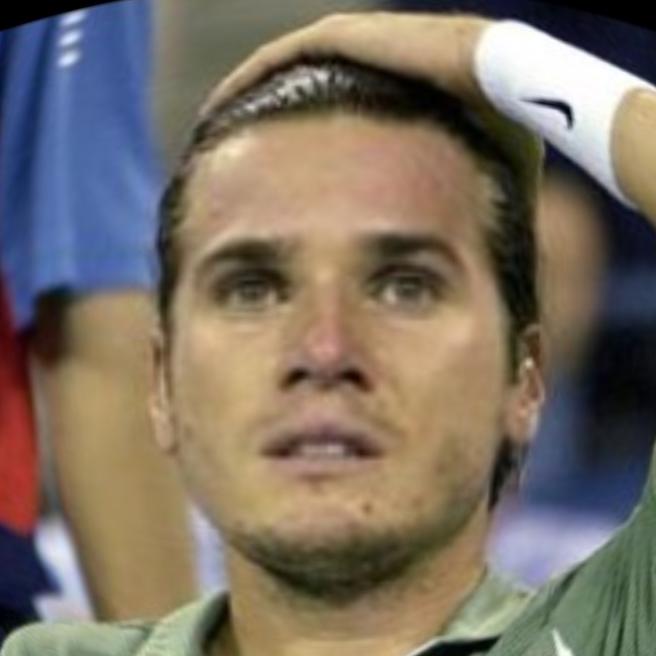}
         \caption{DM\textsubscript{$\lambda = 0.4$}}
         \label{fig:lfw-1-dm4}
     \end{subfigure}
     \hfill
     \begin{subfigure}[b]{0.12\textwidth}
         \centering
         \includegraphics[width=\textwidth]{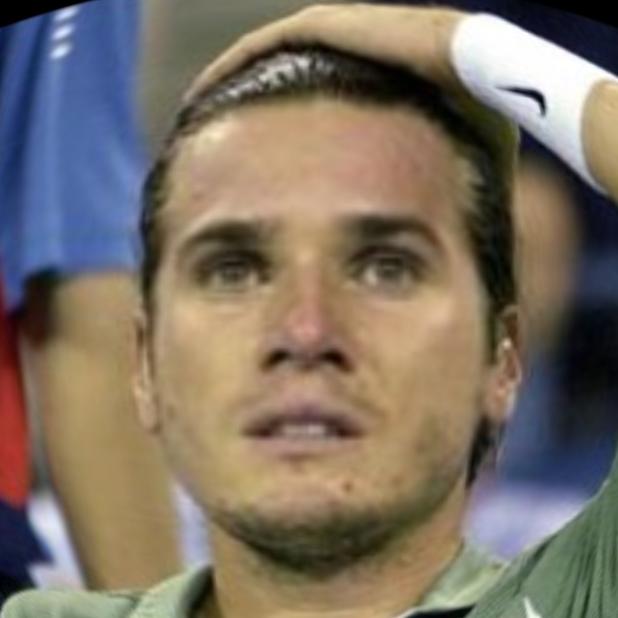}
         \caption{DM\textsubscript{$\lambda = 0.5$}}
         \label{fig:lfw-1-dm5}
     \end{subfigure}
     \hfill
     \begin{subfigure}[b]{0.12\textwidth}
         \centering
         \includegraphics[width=\textwidth]{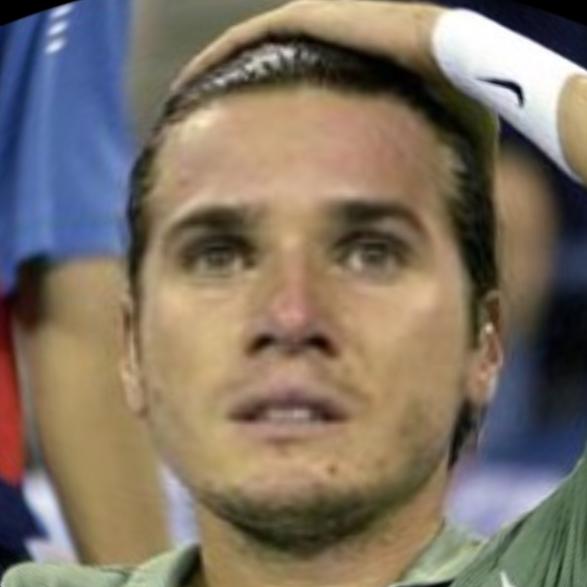}
         \caption{DM\textsubscript{$\lambda = 0.6$}}
         \label{fig:lfw-1-dm6}
     \end{subfigure}
     \hfill
     \begin{subfigure}[b]{0.12\textwidth}
         \centering
         \includegraphics[width=\textwidth]{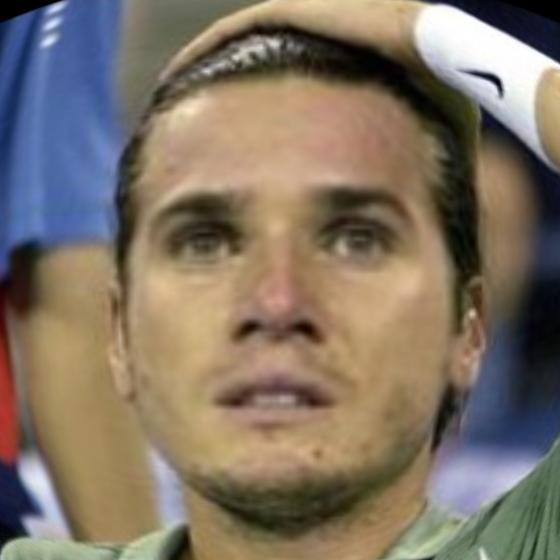}
         \caption{DM\textsubscript{$\lambda = 0.7$}}
         \label{fig:lfw-1-dm7}
     \end{subfigure}
     \hfill
     \begin{subfigure}[b]{0.12\textwidth}
         \centering
         \includegraphics[width=\textwidth]{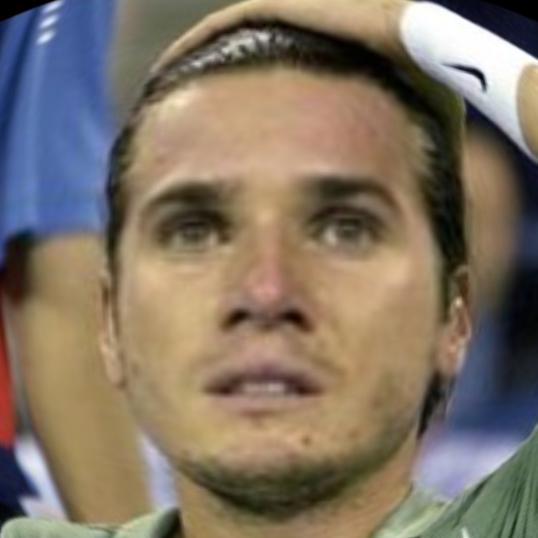}
         \caption{DM\textsubscript{$\lambda = 0.8$}}
         \label{fig:lfw-1-dm8}
     \end{subfigure}
     \hfill
     \begin{subfigure}[b]{0.12\textwidth}
         \centering
         \includegraphics[width=\textwidth]{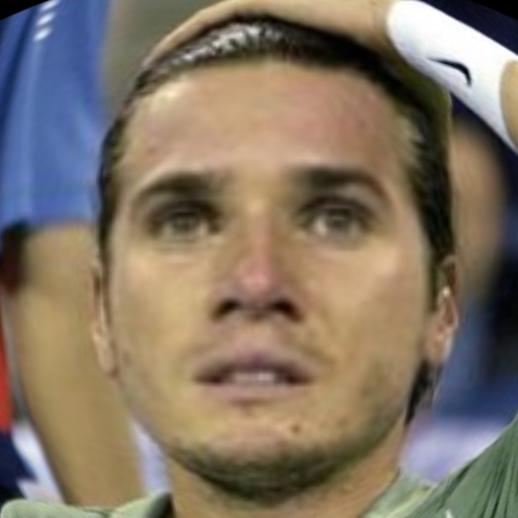}
         \caption{DM\textsubscript{$\lambda = 0.9$}}
         \label{fig:lfw-1-dm9}
     \end{subfigure}
     \hfill
     \begin{subfigure}[b]{0.12\textwidth}
         \centering
         \includegraphics[width=\textwidth]{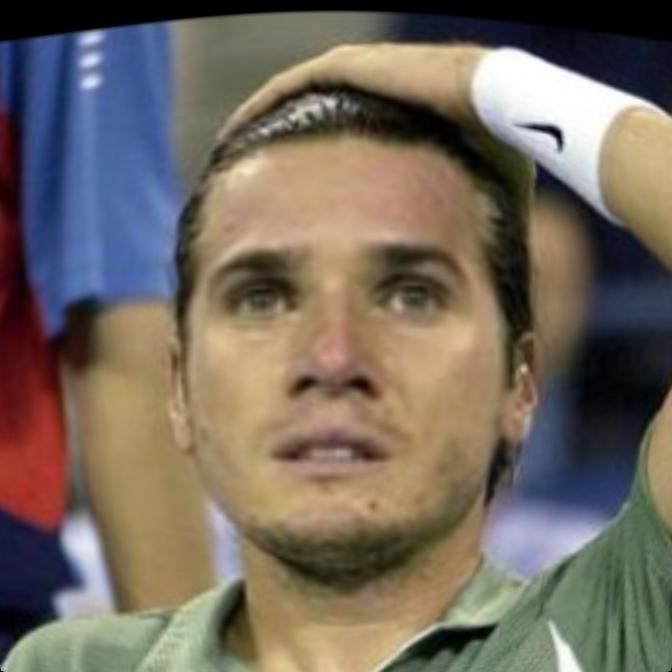}
         \caption{KBp\textsubscript{$\lambda = 1.5$}}
         \label{fig:lfw-1-kbp1.5}
     \end{subfigure}
     \hfill
     \begin{subfigure}[b]{0.12\textwidth}
         \centering
         \includegraphics[width=\textwidth]{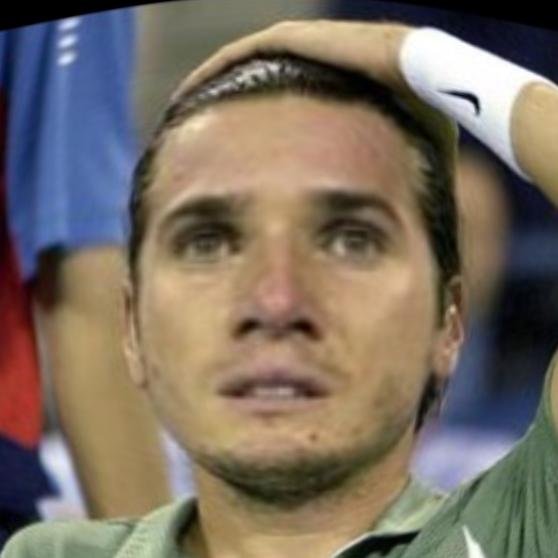}
         \caption{KBp\textsubscript{$\lambda = 2.5$}}
         \label{fig:lfw-1-kbp2.5}
     \end{subfigure}
     \hfill
     \begin{subfigure}[b]{0.12\textwidth}
         \centering
         \includegraphics[width=\textwidth]{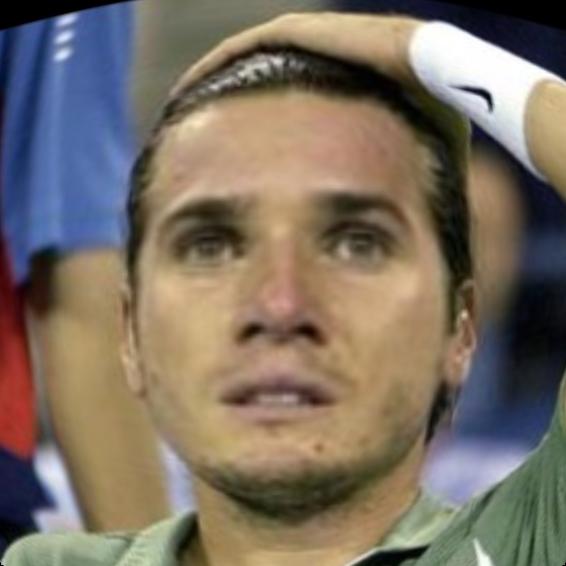}
         \caption{KBs\textsubscript{$\lambda = 1.5$}}
         \label{fig:lfw-1-kbs1.5}
     \end{subfigure}
     \hfill
     \begin{subfigure}[b]{0.12\textwidth}
         \centering
         \includegraphics[width=\textwidth]{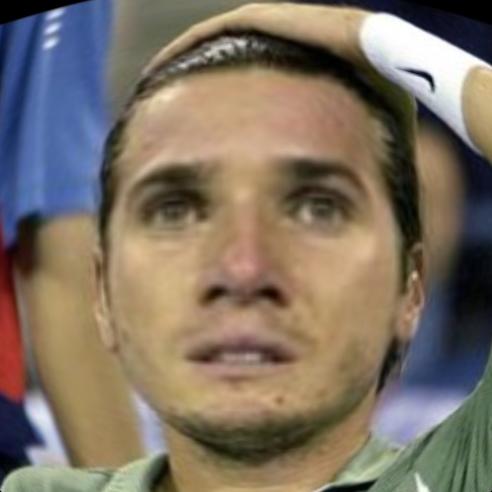}
         \caption{KBs\textsubscript{$\lambda = 2.5$}}
         \label{fig:lfw-1-kbs2.5}
     \end{subfigure}
     \hfill
     \begin{subfigure}[b]{0.12\textwidth}
         \centering
         \includegraphics[width=\textwidth]{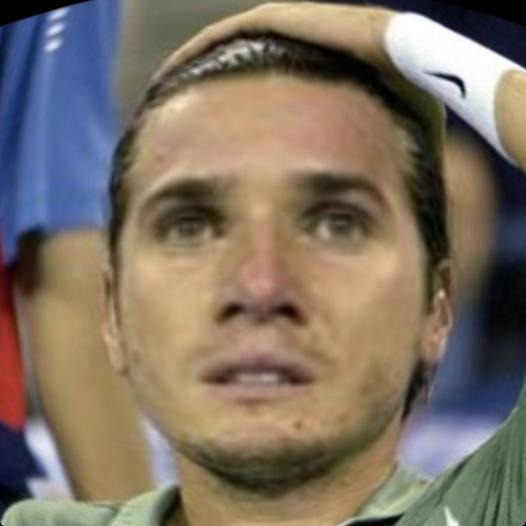}
         \caption{KBe\textsubscript{$\lambda = 1.5$}}
         \label{fig:lfw-1-kbe1.5}
     \end{subfigure}
     \hfill
     \begin{subfigure}[b]{0.12\textwidth}
         \centering
         \includegraphics[width=\textwidth]{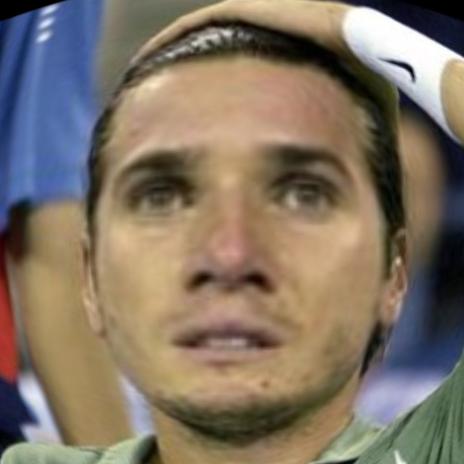}
         \caption{KBe\textsubscript{$\lambda = 2.5$}}
         \label{fig:lfw-1-kbe2.5}
     \end{subfigure}
     \hfill
     \begin{subfigure}[b]{0.12\textwidth}
         \centering
         \includegraphics[width=\textwidth]{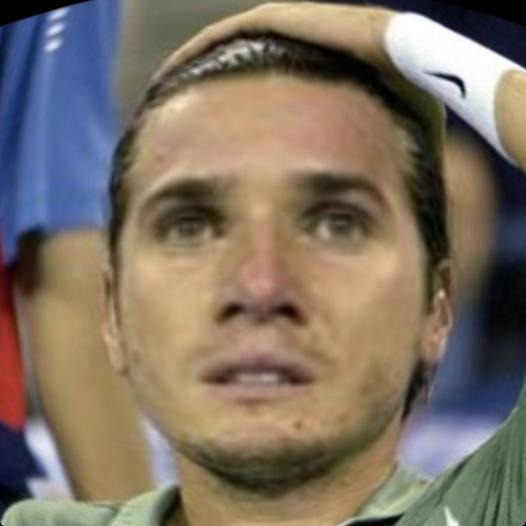}
         \caption{KBo\textsubscript{$\lambda = 1.5$}}
         \label{fig:lfw-1-kbo1.5}
     \end{subfigure}
     \hfill
     \begin{subfigure}[b]{0.12\textwidth}
         \centering
         \includegraphics[width=\textwidth]{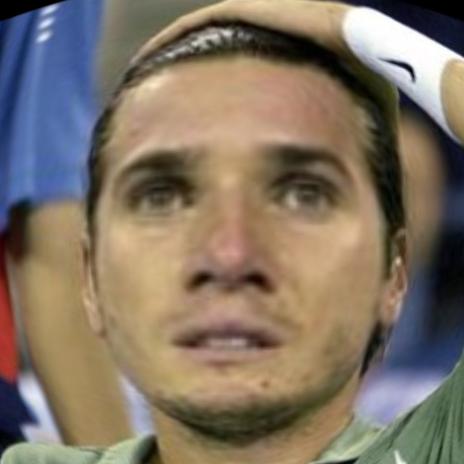}
         \caption{KBo\textsubscript{$\lambda = 2.5$}}
         \label{fig:lfw-1-kbo2.5}
     \end{subfigure}
     \hfill
     \begin{subfigure}[b]{0.12\textwidth}
         \centering
         \includegraphics[width=\textwidth]{source_Tommy_Haas_0002.jpg}
         \caption{Original}
         \label{fig:lfw-1-original1}
     \end{subfigure}
     \end{center}
    \caption{Illustration of the radial distortion effect on a face image using different distortion models and different distortion coefficient values. DM refers to the Division Model. KBp, KBs, KBe, and KBo refer to the Kannala-Brandt model of types perspective, stereographic, equisolid, and orthogonal, respectively. $\lambda$ is the distortion coefficient. Image from LFW dataset \cite{LFWTech}.}
    \label{fig:lfw-distortion-samples}
\end{figure*}

\subsubsection{Kannala-Brandt Model}

The Kannala-Brandt model (KB) defines $\theta$ as the angle between the principal axis and the incoming ray, and it supports different kinds of projections by changing the formula for computing $\theta$. The KB model introduces the general form of the projection, or the $\delta(r_d)$ function, as in Equation \ref{eq:kb-general}, where even powers are dropped \cite{kannala2006generic}, and $\lambda_n$ are the distortion coefficients for a chosen $n$:

\begin{equation}
\delta(r_d) = \lambda_1 \theta + \lambda_2 \theta^3 + \lambda_3 \theta^5 + \lambda_4 \theta^7 + ... 
\label{eq:kb-general}
\end{equation}

The KB is a general model and can express multiple kinds of projections, namely, perspective, stereographic, equidistance, equisolid, and orthogonal. The general form of the projection, illustrated in Equation \ref{eq:kb-general}, is the same for all projection kinds, but the formula for computing $\theta$ is different. Equation \ref{eq:kb-perspective} shows $\theta$ for the perspective projection of a pinhole camera, and Equations \ref{eq:kb-stereographic}, \ref{eq:kb-equidistance}, \ref{eq:kb-equisolid}, and \ref{eq:kb-orthogonal} show $\theta$ for the stereographic, equidistance, equisolid, and orthogonal projections, respectively, which are the ones suitable for modeling fish-eye lenses \cite{kannala2006generic} and where $f$ is the focal length.

\begin{align}
\theta &= \arctan \left( \frac{r_d}{f} \right) & \text{(perspective)} \label{eq:kb-perspective} \\ 
\theta &= 2\arctan \left( \frac{r_d}{2f} \right) & \text{(stereographic)} \label{eq:kb-stereographic} \\
\theta &= \frac{r_d}{f} & \text{(equidistance)} \label{eq:kb-equidistance} \\
\theta &= 2\arcsin \left( \frac{r_d}{2f} \right) & \text{(equisolid)} \label{eq:kb-equisolid} \\
\theta &= \arcsin \left( \frac{r_d}{f} \right) & \text{(orthogonal)} \label{eq:kb-orthogonal}
\end{align}

The relationship between the coordinates $(x_d, y_d)$ of a point in the distorted image, and the coordinates $(x_u, y_u)$ of its corresponding location in the undistorted image is designated as follows:

\begin{equation}
x_u = \frac{\delta(r_d) * x_d}{r_d}
\label{eq:kb-x}
\end{equation}

\begin{equation}
y_u = \frac{\delta(r_d) * y_d}{r_d}
\label{eq:kb-y}
\end{equation}

As in the division model, we use only the first distortion coefficient, and we set the focal length $f$ to $f = 1$. This simplifies equations \ref{eq:kb-x} and \ref{eq:kb-y} to the following:

\begin{equation}
x_u = \frac{\lambda \theta * x_d}{r_d}
\label{eq:kb-x-sim}
\end{equation}

\begin{equation}
y_u = \frac{\lambda \theta * y_d}{r_d}
\label{eq:kb-y-sim}
\end{equation}

Figure \ref{fig:lfw-distortion-samples} illustrates the effect of both distortion models with different configurations and distortion coefficient values on a face image.

\subsection{Radial Distortion Datasets}

The datasets that can be used for training and evaluation must contain distorted and undistorted images. To obtain such datasets, we follow two approaches: (1) we use a set of face image datasets as a basis and a source for the undistorted images, then we use the radial distortion transformations, introduced in Section \ref{sec:trans}, to synthetically create their distorted counterparts; and (2) to evaluate our approach for a real-life scenario, we further collect a dataset with distorted and undistorted images using a mobile phone device.

\subsubsection{Base Datasets}
\label{sec:basedatasets}

We obtain undistorted face images covering three scenarios for face images with EasyPortrait having portrait and selfie images taken by the subjects themselves, FRLL having images taken in a studio setup, and LFW having images taken in the wild. The EasyPortrait (EP) dataset \cite{kapitanov2023easyportrait} is a large-scale image dataset for portrait segmentation and face parsing. The dataset offers a diverse range of real-life scenarios in which people are taking selfies or portrait photos in front of a laptop or a smartphone camera. It contains, in total, 20,000 indoor photos of 8,377 distinct subjects. The dataset is split into three subsets: training, validation, and testing. These subsets contain 14000, 2000, and 4000 images, respectively. The Face Research Lab London Set (FRLL)  \cite{DeBruine2021} contains 1020 images of 102 subjects. The images are taken in a studio setup under the same environmental conditions, such as lightning, camera, and camera-to-subject distance. It features five pose variations per subject (frontal, left profile, right profile, left 3 quarter, right 3 quarter), and two expression variations (neutral and smiling). We use only the frontal images of each subject in this work because we want the radial distortion to be the major factor of variation when analyzing the impact on face recognition. The Labeled Faces in the Wild (LFW) dataset \cite{LFWTech} contains face photographs with more than 13,000 images of faces collected from the web in an unconstrained setting. The total number of subjects is 5,750, of which 1680 have two or more distinct photos.

\subsubsection{Synthetic Datasets}
\label{sec:synthetic}

The transformation models introduced in Section \ref{sec:trans} can be used to transform a distorted image into a non-distorted image and vice versa. Using these models, we generate distorted counterparts for the images in the base datasets. With three base datasets and two camera models, and a distortion coefficient that can take a range of different values, we create synthetic datasets that vary across three dimensions: (1) the base dataset, (2) the camera model used to produce the radially distorted images, and (3) the value chosen for the distortion coefficient (which governs the intensity of the distortion). Table \ref{tbl:datasets} describes all the synthetic datasets created for the purposes of training and evaluating the models in this work. It lists the name given to each dataset, its corresponding base dataset used to obtain the undistorted images, the distortion model (and the specific variant in the case of KB model) used to introduce the distortion effect, the value of the distortion coefficient $\lambda$, and whether the dataset is used for training, evaluation, or both.

\begin{table}[h]
\begin{center}
\begin{tabular}{lllll}
\hline
\hline
Name & Base & Model & $\lambda$ & T/E \\ 
\hline
\hline
$EP_{dm/0.3}$ & EP & DM & 0.3  & T \\ 
$EP_{dm/0.6}$ & EP & DM & 0.6  & T \\ 
$EP_{dm/0.9}$ & EP & DM & 0.9  & T\&E \\ 
$FRLL_{dm/0.6}$ & FRLL & DM & 0.6  & T \\ 
$LFW_{dm/0.6}$  & LFW  & DM & 0.6  & T \\ 
$EP_{dm/0.4}$ & EP & DM & 0.4  & E \\ 
$FRLL_{dm/0.4}$ & FRLL & DM & 0.4  & E \\ 
$FRLL_{kbp/1.5}$ & FRLL & KBp & 1.5  & E \\ 
$LFW_{dm/0.4}$ & LFW  & DM & 0.4  & E \\ 
$LFW_{dm/0.7}$ & LFW  & DM & 0.7  & E \\ 
$LFW_{kbe/1.5}$ & LFW  & KBe & 1.5  & E \\ 
\hline
\hline
\end{tabular}
\end{center}
\caption{The synthetic datasets created for the purposes of training and evaluating the models. "Base" refers to one of the base datasets described in Section \ref{sec:basedatasets}. "Model" refers to the distortion model used to create the dataset, where "DM" is the devision model, "KBp" is the perspective variant of the Kannala-Brandt model, and "KBe" is the equisolid one. $\lambda$ is the distortion coefficient. "T/E" refers to whether the dataset is employed for training (T) or evaluation (E) or both (T\&E).}
\label{tbl:datasets}
\end{table}

\subsubsection{Mobile Fisheye Dataset}
\label{sec:mfd}

\begin{figure}[h]
     \begin{center}
     \resizebox{0.6\columnwidth}{!}{%
     \begin{subfigure}[b]{0.45\linewidth}
         \centering
         \includegraphics[width=\linewidth]{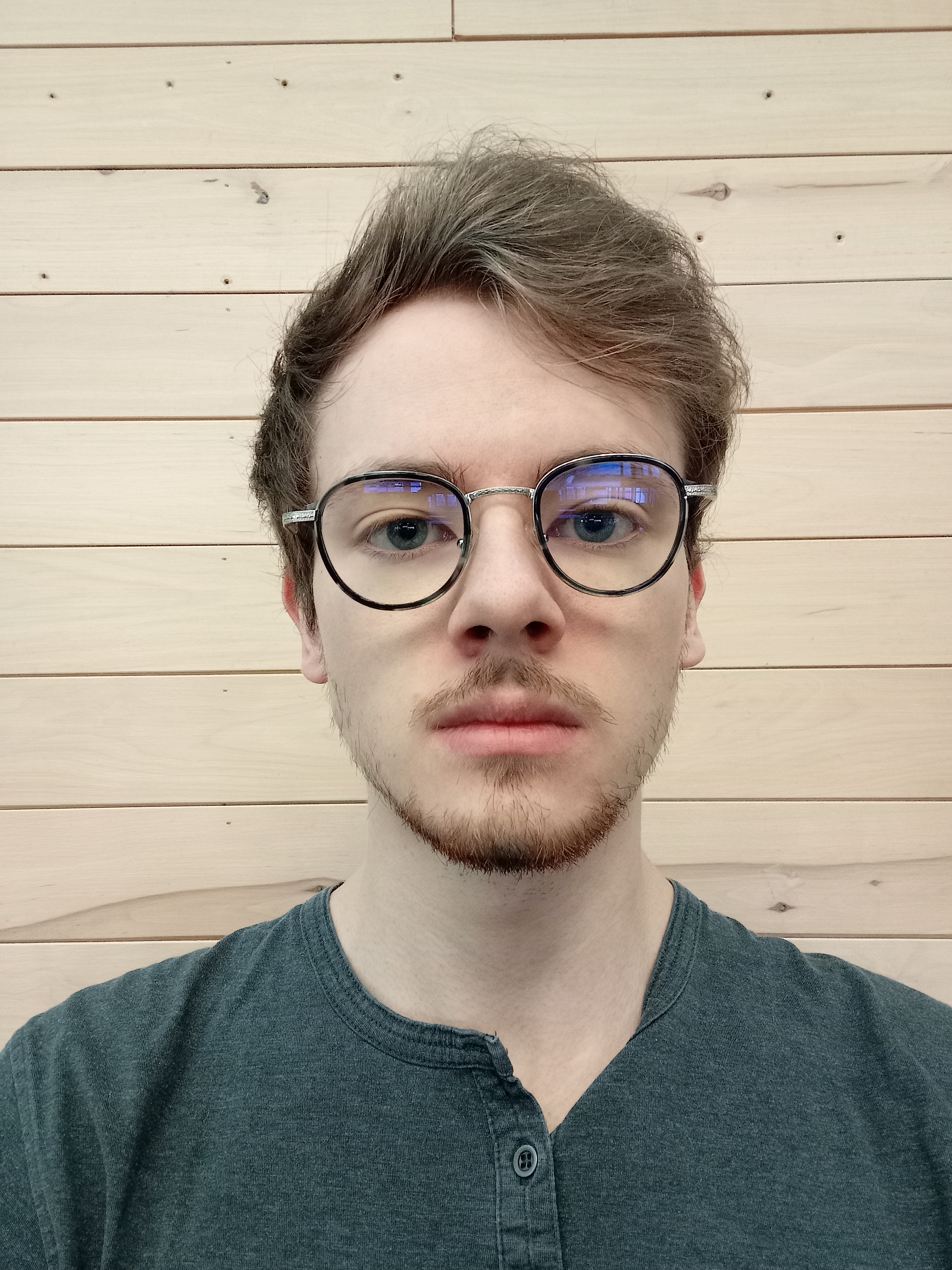}
         \caption{Undistorted}
         \label{fig:mfd-undistorted}
     \end{subfigure}
    \hfill
     \begin{subfigure}[b]{0.45\linewidth}
         \centering
         \includegraphics[width=\linewidth]{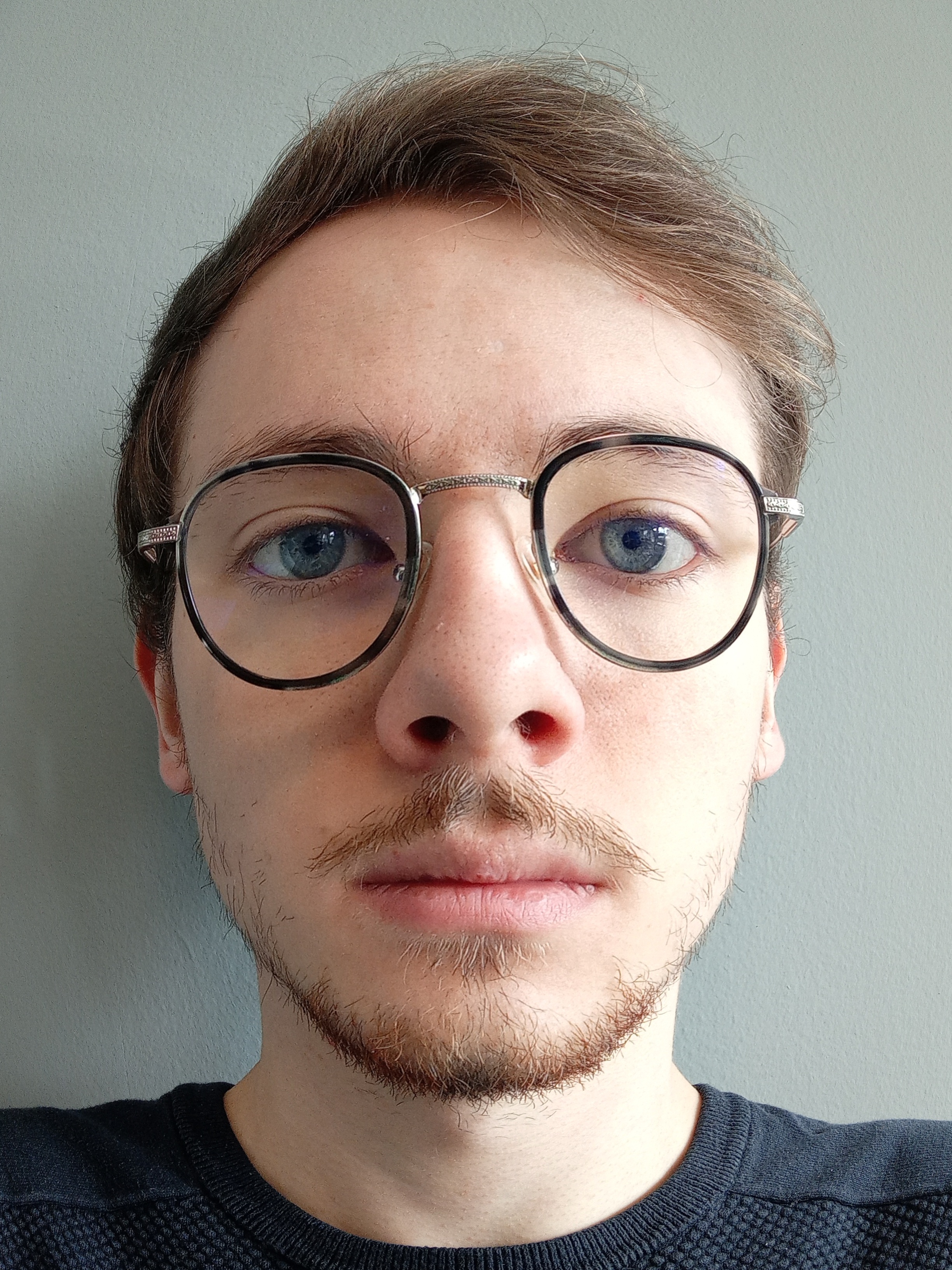}
         \caption{Distorted}
         \label{fig:mfd-distorted}
     \end{subfigure}
     }
     \end{center}
    \setlength{\belowcaptionskip}{-8pt}
    \caption{Sample distorted and undistorted images from the collected Mobile Fisheye Dataset (MFD)}
    \label{fig:mfd}
\end{figure}

Mobile Fisheye Dataset collected in this work uses a mobile phone device and an off-the-shelf camera app that takes wide-angle photos. The app allows the distortion level to be manipulated by the user, so the distorted images are produced with random levels of distortion. Table \ref{tbl:mfd-dataset} provides an overview of the dataset and Figure \ref{fig:mfd} shows sample distorted and undistorted images. We call this dataset the Mobile Fisheye Dataset (MFD).

\begin{table}[h]
\begin{center}
\begin{tabular}{cccc}
\hline
\hline
\#Images & \#Subjects & \#Distorted & \#Undistorted \\ 
\hline
\hline
188 & 5 & 130 & 58 \\
\hline
\hline
\end{tabular}
\end{center}
\caption{Overview of the MFD dataset.}
\label{tbl:mfd-dataset}
\end{table}

\subsection{Radial Distortion Detection}

The radial distortion detector is a simple binary classification model. The model is a deep neural network based on a pre-trained ResNet34 model \cite{he2016deep}. Using transfer learning, we fine-tune the model on the chosen training dataset. The model originally had a 1000-dimensional output layer, but our dataset has only 2 classes (distorted and undistorted), so we removed the output layer and defined a new fully connected layer with just 2 neurons, one for each class. Since the primary purpose of the detection model is to be able to flag distorted images, particularly selfies and portraits, that can be taken by a mobile phone camera and provided to an ID issuance system, the dataset that fits this purpose best is the \textit{EasyPortrait} dataset. Thus, it is the one chosen as the primary training dataset. In particular, the training subset of the dataset is used for training, and the other two subsets are used for model validation and testing. Then, to make sure that the model is evaluated against unseen distortion models, we choose to limit the training to datasets distorted with the Division Model (DM) only and use datasets created with the Division Model (DM) and the Kannala-Brandt (KB) model, with any of its variants, for evaluation.

\subsection{Radial Distortion as FIQA Algorithm}
\label{sec:approach-fiqa}

To study the effect of radial distortion on the FRS performance, we formulate the radial distortion detection model as a face image quality assessment (FIQA) algorithm that produces a numerical output that monotonically increases as the quality of the face image increases, leading to an increased utility of the image for a FRS. To produce this numerical output, which we call from here on \textit{native quality measure (NQM)} to align with the terminology used in the ISO/IEC 29794-5 standard \cite{ISO-IEC-29794-5-DIS-FaceQuality-240129}, from the detection model, we use the normalized outputs of the last layer of the network. More formally, given that $\alpha$ and $\beta$ are the two raw output logits of the network for classes \textit{distored} and \textit{undistorted}, respectively, we consider the native quality measure to be the following:

\begin{equation}
NQM = softmax(\beta) = \frac{e^{\beta}}{e^{\alpha} + e^{\beta}}
\label{eq:quality-score}
\end{equation}

This acts as an indicator of how confident the model is about the image being undistorted and can be used as a native quality measure.

\section{Experiments}
\label{sec:experiments}

\subsection{Radial Distortion Detection Models}

\begin{figure*}[htp]
     \begin{center}
     \begin{subfigure}[b]{0.24\textwidth}
         \centering
         \includegraphics[width=\textwidth]{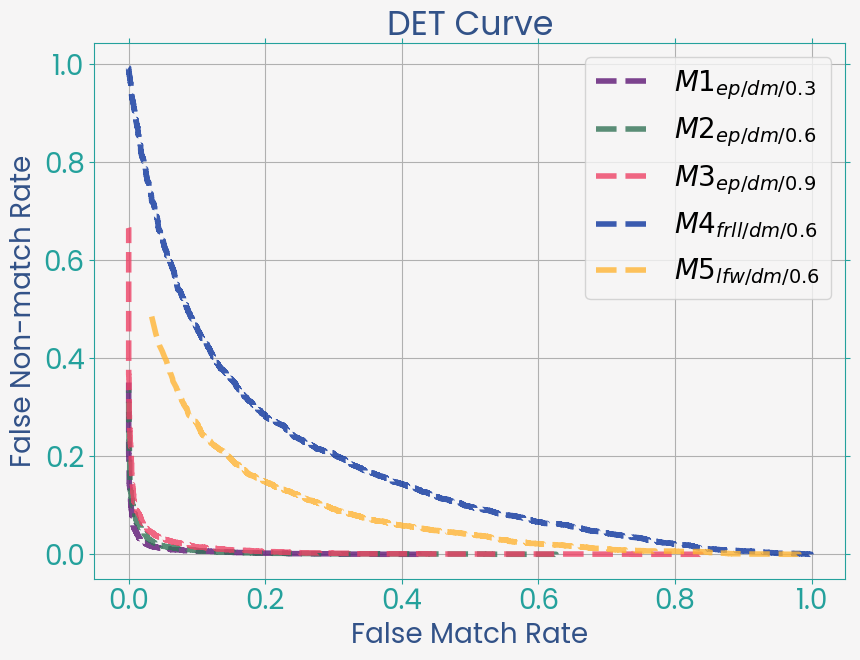}
         \caption{$EP_{dm/0.4}$}
         \label{fig:eval-ep-dm-0.4}
     \end{subfigure}
     \hfill
     \begin{subfigure}[b]{0.24\textwidth}
         \centering
         \includegraphics[width=\textwidth]{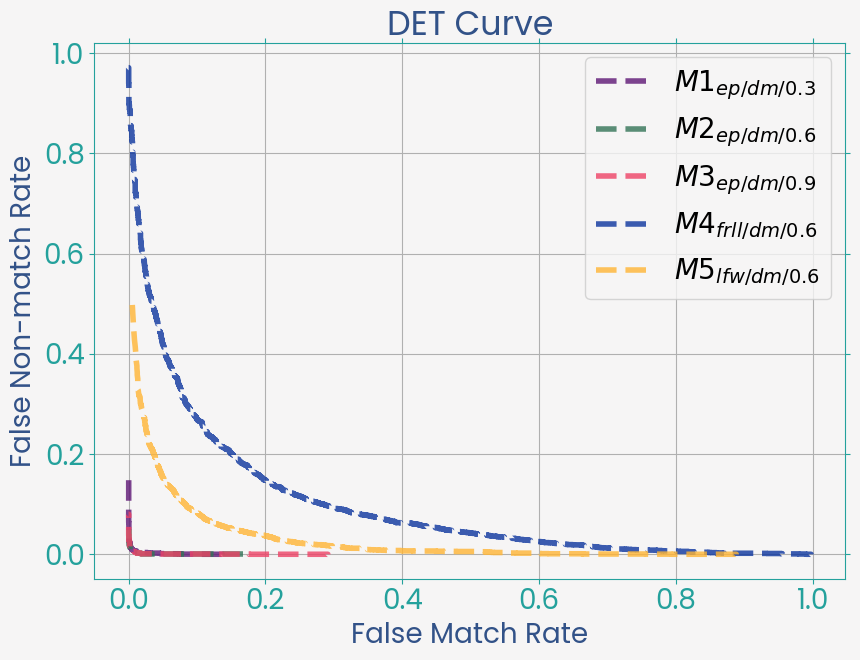}
         \caption{$EP_{dm/0.9}$}
         \label{fig:eval-ep-dm-0.9}
     \end{subfigure}
     \hfill
     \begin{subfigure}[b]{0.24\textwidth}
         \centering
         \includegraphics[width=\textwidth]{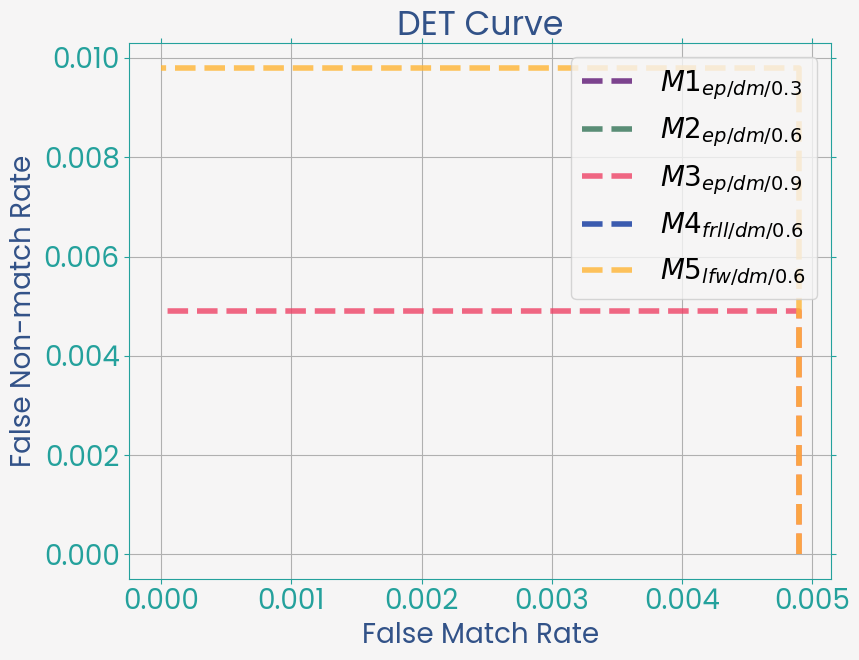}
         \caption{$FRLL_{dm/0.4}$}
         \label{fig:eval-frll-dm-0.4}
     \end{subfigure}
     \hfill
     \begin{subfigure}[b]{0.24\textwidth}
         \centering
         \includegraphics[width=\textwidth]{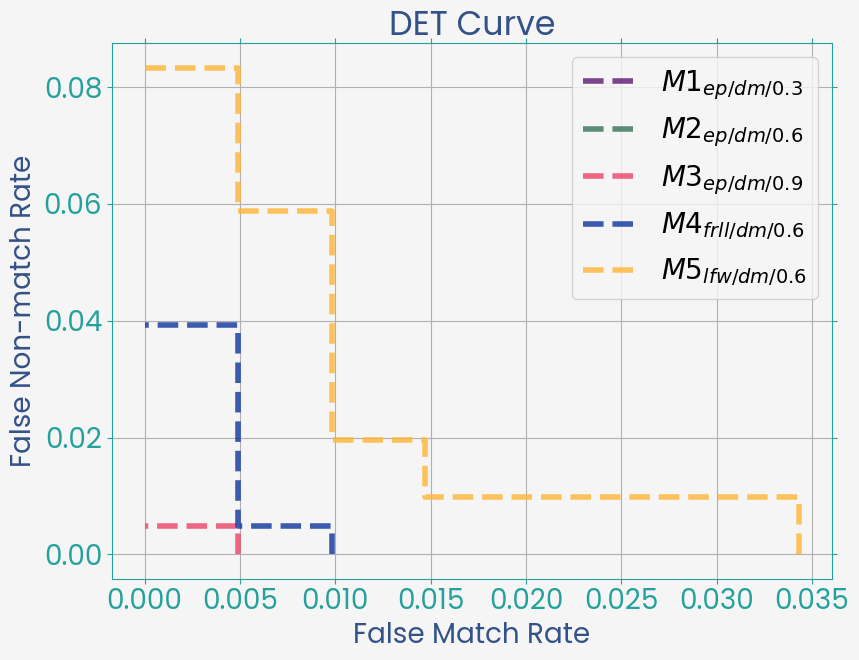}
         \caption{$FRLL_{kbp/1.5}$}
         \label{fig:eval-frll-kbp-1.5}
    \end{subfigure}
    \hfill
    \begin{subfigure}[b]{0.24\textwidth}
         \centering
         \includegraphics[width=\textwidth]{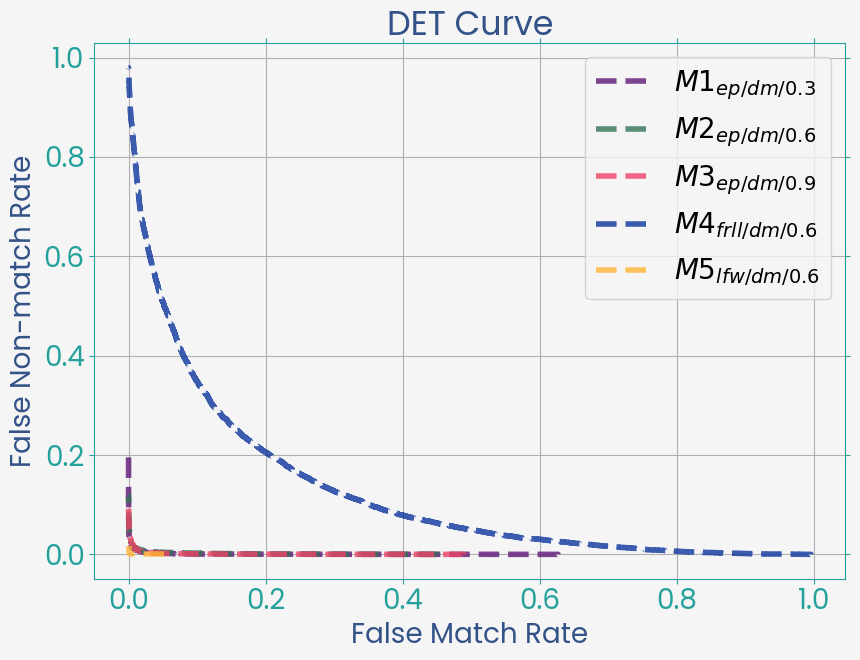}
         \caption{$LFW_{dm/0.4}$}
         \label{fig:eval-lfw-dm-0.4}
     \end{subfigure}
     \hfill
     \begin{subfigure}[b]{0.24\textwidth}
         \centering
         \includegraphics[width=\textwidth]{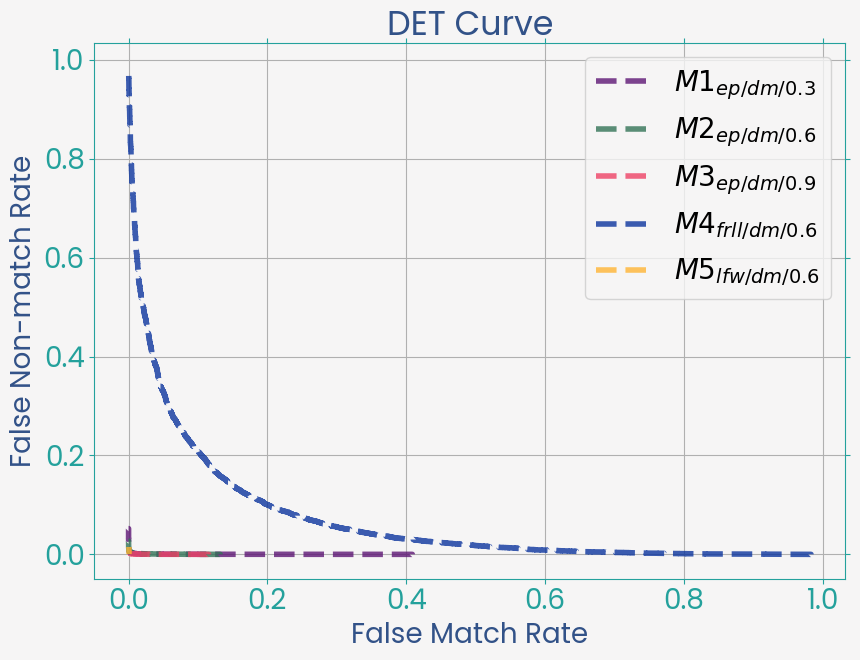}
         \caption{$LFW_{dm/0.7}$}
         \label{fig:eval-lfw-dm-0.7}
     \end{subfigure}
     \hfill
     \begin{subfigure}[b]{0.24\textwidth}
         \centering
         \includegraphics[width=\textwidth]{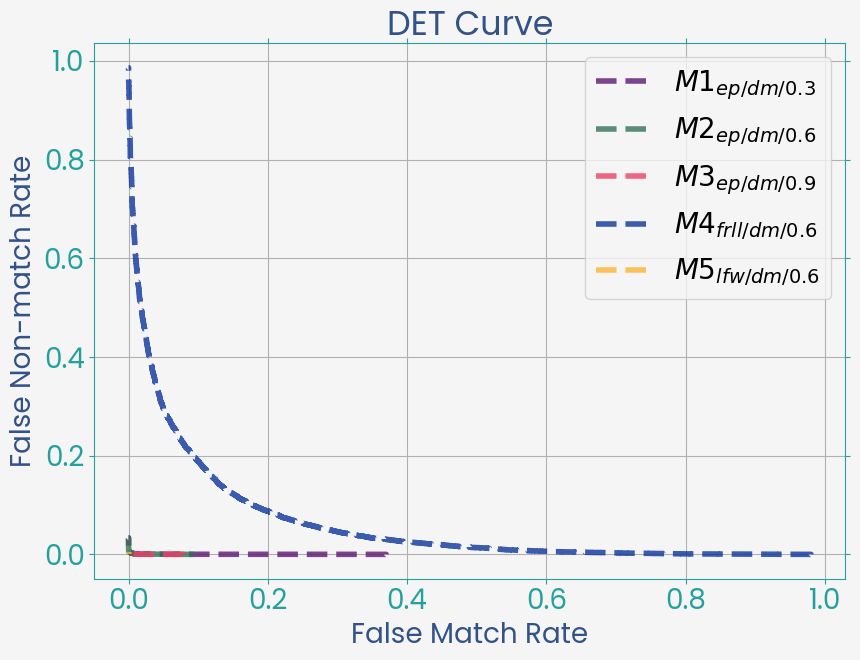}
         \caption{$LFW_{kbe/1.5}$}
         \label{fig:eval-lfw-kbe-1.5}
     \end{subfigure}
     \hfill
     \begin{subfigure}[b]{0.24\textwidth}
         \centering
         \includegraphics[width=\textwidth]{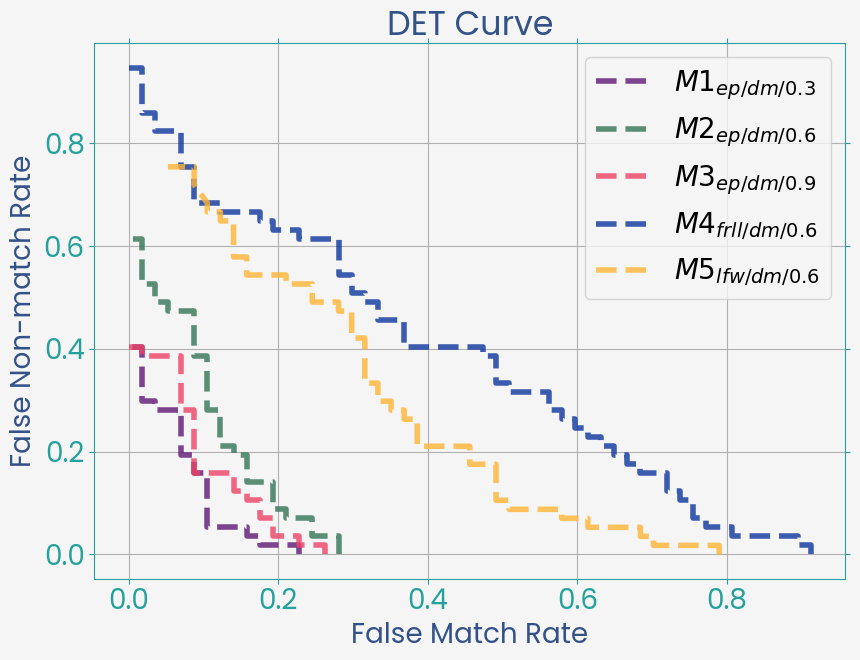}
         \caption{$MFD$}
         \label{fig:eval-mfd}
     \end{subfigure}
     \end{center}
    \caption{Evaluation results for the five distortion detection models on datasets with different base datasets, distortion models, and distortion coefficient values. Figure~\ref{fig:eval-ep-dm-0.4} to \ref{fig:eval-lfw-kbe-1.5} show evaluation results on the synthetic datasets. Figure \ref{fig:eval-mfd} shows evaluation results on the self-collected Mobile Fisheye Dataset (MFD).}
    \label{fig:det}
\end{figure*}

To examine which training dataset and distortion coefficient value give the best classification results, we train five models on synthetic datasets as described in Table \ref{tbl:models}. Details about how each dataset was created are listed in Table \ref{tbl:datasets}. Each model is identified by a number and the dataset it is trained on. For example, model $M1_{EP/dm/0.3}$ is trained on the synthetic dataset $EP_{dm/0.3}$, whose details are described in Table \ref{tbl:datasets}. To make sure that each model is evaluated across datasets created with different distortion models and distortion coefficient values, the evaluation of each of the five models is performed on multiple synthetic datasets as well as the MFD. The Detection Error Tradeoff (DET) curves, shown in Figure \ref{fig:det}, are used to report the evaluation results. Figure \ref{fig:eval-ep-dm-0.4} shows the evaluation results on the $EP_{dm/0.4}$ dataset created based on the EasyPortrait dataset, using the division model, and with a relatively mild distortion with $\lambda=0.4$. The DET curves clearly show that models $M1$, $M2$, and $M3$ have rather identical and perfect results. Models $M4$ and $M5$ have lower performance, which is understandable given that they are trained on the $FRLL_{dm/0.6}$ and $LFW_{dm/0.6}$ datasets, which contain fewer training images compared to the $EP$ variants. For brevity, Figures \ref{fig:eval-ep-dm-0.9} through \ref{fig:eval-lfw-kbe-1.5} will not be described in detail as they show rather the same results as the ones explained in the previous figure, with models $M1$, $M2$, and $M3$ having rather identical and perfect results. Note that Figures \ref{fig:eval-frll-dm-0.4} and \ref{fig:eval-frll-kbp-1.5} have different X and Y scales because the error rate is very small on all models and the scale is thus adjusted to accommodate this in the plot. Figure \ref{fig:eval-mfd} shows the evaluation results on the MFD. The curves clearly show that models $M1$, $M2$, and $M3$ are still performing very well over a dataset with distorted images produced by an unknown method in a mobile app. The overall evaluation results show that all three models, $M1$, $M2$, and $M3$ are performing the best. They generalize well across other datasets, distortion models, and distortion coefficient values. They also perform well over a dataset of images produced on a mobile device with an unknown distortion algorithm, distortion coefficients, and camera specifications.

\begin{table}[h]
\begin{center}
\begin{tabular}{ll}
\hline
\hline
Model Name & Training Dataset \\ 
\hline
\hline
$M1_{EP/dm/0.3}$ & $EP_{dm/0.3}$ \\ 
$M2_{EP/dm/0.6}$ & $EP_{dm/0.6}$  \\ 
$M3_{EP/dm/0.9}$ & $EP_{dm/0.9}$  \\ 
$M4_{FRLL/dm/0.6}$ & $FRLL_{dm/0.6}$  \\ 
$M5_{LFW/dm/0.6}$ & $LFW_{dm/0.6}$  \\ 
\hline
\end{tabular}
\end{center}
\caption{The names given to the five radial distortion detection models and the name of the dataset each model is trained on.}
\label{tbl:models}
\end{table}

\subsection{Radial Distortion and Face Recognition}

\begin{figure}[h]
     \begin{center}
     \begin{subfigure}[b]{0.49\linewidth}
         \centering
         \includegraphics[width=\linewidth]{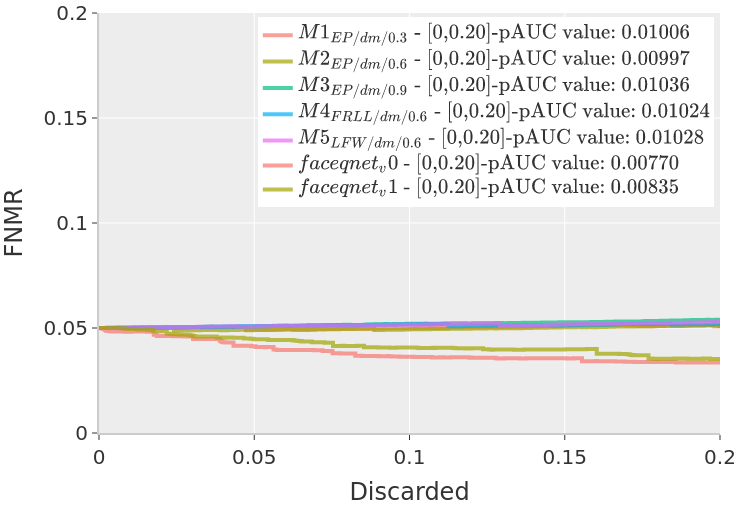}
         \caption{$LFW_{dm/0.1-0.9}$}
         \label{fig:edc-lfw-dm-0.1-0.9}
     \end{subfigure}
    \hfill
     \begin{subfigure}[b]{0.49\linewidth}
         \centering
         \includegraphics[width=\linewidth]{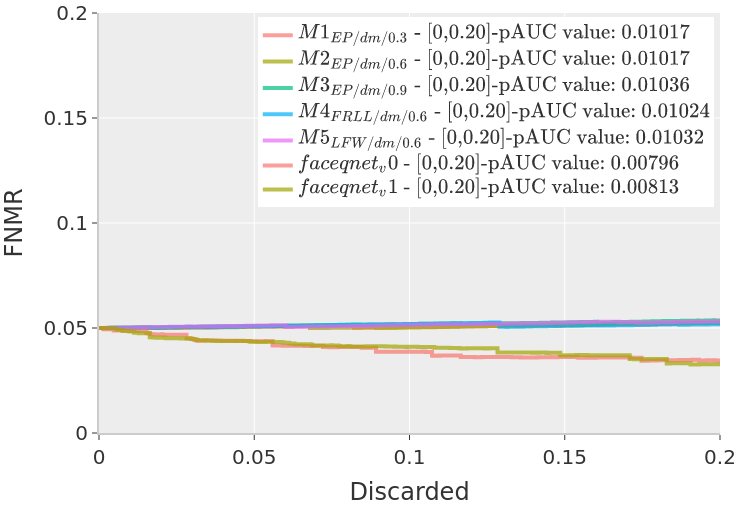}
         \caption{$LFW_{kbs/1.0-2.5}$}
         \label{fig:edc-lfw-kbs-1.0-2.5}
     \end{subfigure}
     \end{center}
    \caption{EDC curves on two synthetically distorted versions of the LFW dataset, created with random distortion levels ranging from very mild to very severe distortion. ArcFace is used as the FRS, and 0.05 is the starting error rate.}
    \label{fig:edc}
\end{figure}

To evaluate the effect of radial distortion on face recognition, we use the "Error versus Discard Characteristic" (EDC) curves. The EDC curve is standardized in the international standard ISO/IEC 29794-1 \cite{ISO-IEC-29794-1-QualityFramework-231013}. The EDC curve illustrates the correlation between the discard percentage of the lower quality images, based on a quality measure produced by a FIQA algorithm, and an error rate, herein False Non-Match Rate (FNMR). We use two synthetically distorted variants of the LFW dataset for the evaluation since the original LFW dataset does not contain radially distorted images. The first one, $LFW_{dm/0.1-0.9}$, is created using the Division Model (DM) with a distortion coefficient $\lambda \in [0.1, 0.9]$. The second one, $LFW_{kbs/1.0-2.5}$, is created using the KB model with the stereographic variant and a distortion coefficient $\lambda \in [1.0, 2.5]$. We produce EDC curves for the five models using the quality measures produced as described in Section \ref{sec:approach-fiqa} along with two reference FIQA algorithms, namely \textit{FaceQNet V0} and \textit{FaceQNet V1}. These algorithms produce a unified quality score, which is an overall assessment of the face image, in other words, they do not specifically measure radial distortion or any other individual defect of the image. The EDC curves are produced using ArcFace \cite{deng2019arcface} as the face recognition system and the framework created by Schlett \etal \cite{schlett2023considerations}. The results are shown in Figure \ref{fig:edc}. As can be seen in the two EDC plots, an impact on the FRS performance is hardly noticeable. The error rate for the five models remains rather constant as a larger percentage of images are discarded. The error rate for the two reference algorithms descends slightly, but it is not considerably different from the ones in the five models. This might lead us to believe that FRS—or at least the ArcFace model—is insensitive to the defect of radial distortion. However, we cannot really conclude this yet. In Section \ref{sec:discussion}, we discuss how the way the EDC curve is produced and the way the FRS pipeline works prevent us from readily drawing such a conclusion.

\section{Discussion}
\label{sec:discussion}

\begin{figure}[htp]
     \begin{center}
     \begin{subfigure}[b]{0.30\linewidth}
         \centering
         \includegraphics[width=\linewidth]{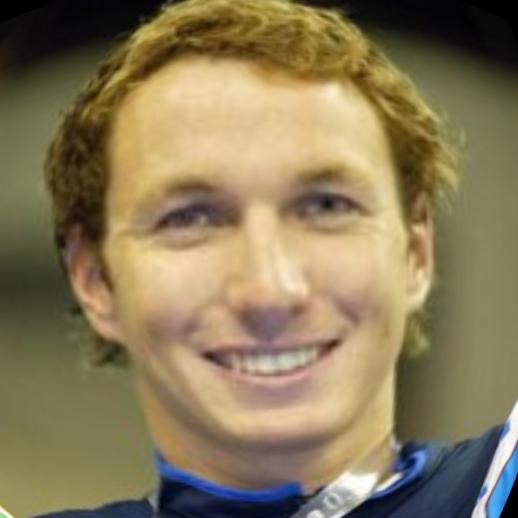}
         \caption{$dm/0.5$}
         \label{fig:pipe-dm-0.5-1}
     \end{subfigure}
     \hfill
     \begin{subfigure}[b]{0.30\linewidth}
         \centering
         \includegraphics[width=\linewidth]{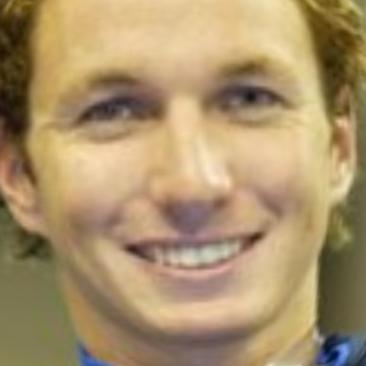}
         \caption{$dm/0.5 \rightarrow fc$}
         \label{fig:pipe-dm-0.5-after-1}
     \end{subfigure}
     \hfill
     \begin{subfigure}[b]{0.30\linewidth}
         \centering
         \includegraphics[width=\linewidth]{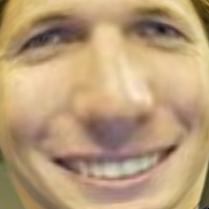}
         \caption{$fc \rightarrow dm/0.5$}
         \label{fig:pipe-dm-0.5-before-1}
     \end{subfigure}
    \hfill
     \begin{subfigure}[b]{0.30\linewidth}
         \centering
         \includegraphics[width=\linewidth]{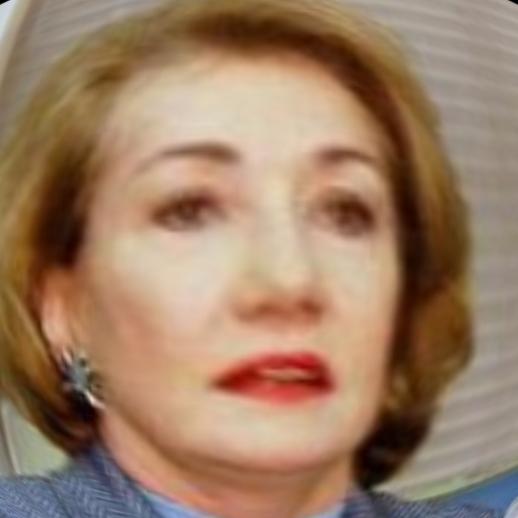}
         \caption{$dm/0.5$}
         \label{fig:pipe-dm-0.5-2}
     \end{subfigure}
     \hfill
     \begin{subfigure}[b]{0.30\linewidth}
         \centering
         \includegraphics[width=\linewidth]{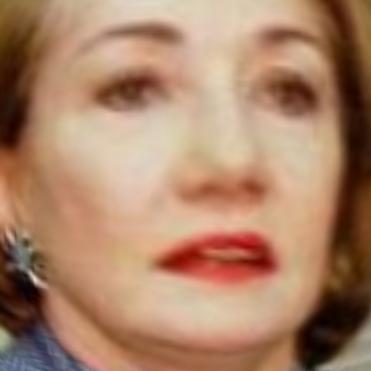}
         \caption{$dm/0.5 \rightarrow fc$}
         \label{fig:pipe-dm-0.5-after-2}
     \end{subfigure}
     \hfill
     \begin{subfigure}[b]{0.30\linewidth}
         \centering
         \includegraphics[width=\linewidth]{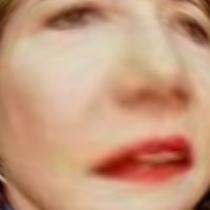}
         \caption{$fc \rightarrow dm/0.5$}
         \label{fig:pipe-dm-0.5-before-2}
     \end{subfigure}
     \end{center}
    \caption{Illustrating the effect of face cropping on radial distortion. The left column shows images distorted with the Division Model (DM) and a distortion coefficient $\lambda=0.5$. The middle column illustrates the effect when the distortion is applied first, then the face is aligned and cropped. The right column illustrates the effect when the images have been aligned and cropped first, then distorted. The difference is quite obvious where, in the middle column, the radial distortion effect becomes less visible and in the right column, the radial distortion becomes more pronounced.}
    \label{fig:pipe}
\end{figure}

To hypothesize about the reason why radial distortion does not have a significant impact on FRS performance, we need a deeper understanding of the pipeline that produces the EDC curve and the stage at which the FIQA process takes place. The FRS pipeline typically includes the following major stages: (1) face detection, (2) face alignment and cropping, (3) feature extraction, and (4) feature comparison. The EDC pipeline naturally resembles the FRS pipeline for the EDC curve to be predictive of FRS performance. Under this framework, the FIQA process happens after stage (2) and before stage (3), i.e., after face cropping and before feature extraction. Figure \ref{fig:pipe} shows sample images of two subjects to illustrate the effect of face cropping on radial distortion. The left column shows the images after being distorted using the Division Model with a distortion coefficient $\lambda=0.5$. The distortion effect is intermediate but noticeable. The middle column shows images that have been first distorted using the same distortion configurations, then face cropping took place. The right column shows images that have been cropped first and then distorted with the same distortion configurations. It is evident from both examples that when the distortion takes place after face alignment and cropping, the distortion effect is much more pronounced and visible (the right column). On the contrary, when the distortion takes place before alignment and cropping, then the distortion is reduced and less noticeable even for the human eye (the middle column). Consequently, it will not be expected to have much of an impact on FRS performance. In the conducted experiments and following the standard setup for producing EDC curves, the distorted datasets are prepared prior to computing the EDC curve, which means distortion is applied first. Then, as the EDC curves are produced, the images will go through face alignment and cropping, which will reduce the distortion effect. This is important because it reflects what takes place in a real-world operational scenario. In practice, a subject might upload a distorted image as part of an ID document application process; this image will run through the FRS pipeline, during which it will be aligned and cropped, which will decrease the radial distortion effect. The image could still be useful for recognition, as shown here, but since we want to collect high-quality and canonical images in our reference databases, the image must not be accepted. This indicates the need for detecting radial distortion earlier in the processing pipeline to flag distorted images. In our primary use case of the unsupervised mobile enrolment scenario, this means that the radial distortion check should be applied before any pre-processing is applied to the image.

\section{Conclusion}
\label{sec:conclusion}

In this paper, we explore a less studied area in the literature, namely radial distortion in face images. We introduce models that can effectively detect distortion during the unsupervised self-enrolment process. We formulate the detection model as a FIQA algorithm that can be used in the standard ISO/IEC 29794-5 \cite{ISO-IEC-29794-5-DIS-FaceQuality-240129} and present a study on the impact of radial distortion on FRS performance. The study reveals that it is better to run such an algorithm before any preprocessing is applied to the image. Future work could make the FIQA algorithm more sensitive to distortion intensity and simplify the detection model even further.

\section*{Acknowledgement}
This work was supported by the European Union’s Horizon 2020 Research and Innovation Program under Grant 883356.


\end{document}